# Abstraction Model for Semantic Segmentation Algorithms

Reihaneh Teymoori[1], Zahra Nabizadeh[2], Nader Karimi[2], Shadrokh Samavi[3]

[1]Department of Computer Science and Software Engineering, Concordia University, Canada
[2]Department of Electrical and Computer Engineering, Isfahan University of Technology, Isfahan, Iran
[3]Department of Electrical and Computer Engineering, McMaster University, Canada

*Abstract*— **Semantic segmentation classifies each pixel in the image. Due to its advantages, semantic segmentation is used in many tasks, such as cancer detection, robot-assisted surgery, satellite image analysis, and self-driving cars. Accuracy and efficiency are the two crucial goals for this purpose, and several state-of-the-art neural networks exist. By employing different techniques, new solutions have been presented in each method to increase efficiency and accuracy and reduce costs. However, the diversity of the implemented approaches for semantic segmentation makes it difficult for researchers to achieve a comprehensive view of the field. In this paper, an abstraction model for semantic segmentation offers a comprehensive view of the field. The proposed framework consists of four general blocks that cover the operation of the majority of semantic segmentation methods. We also compare different approaches and analyze each of the four abstraction blocks' importance in each method's operation.**

## I. Introduction

In semantic segmentation, we try to assign every pixel to different classes [1], [2], [3], [4], and [5]. Hence, in this task, the input image transforms into a pixel-wise classified image [6-8]. In this algorithm, at first, networks try to extract features, and then the pixels in images are classified. There are several applications for semantic segmentation. For example, it has been employed in image retargeting [7], cancer diagnosis and prognosis [8], [9], robot-assisted surgery [10], satellite images [11], traffic management [12], road monitoring [13], face semantic segmentation [14], categorizing clothing items and fashion [15] and farming robots [16]. Different metrics, accuracy, time, memory, and efficiency are essential in these applications. Due to this, researchers have proposed different semantic segmentation methods and tried to improve one of these metrics. In the following, approaches that use neural networks will be explained.

Convolutional networks [17] were a real revolution in object detection. Fully Convolutional Networks (FCN) were the state-of-the-art approach in semantic segmentation. This network takes an image with an arbitrary size as input and produces output in the input size. FCN is used in AlexNet [18], VGG net [19], and GoogLeNet [20]. Besides its advantages, FCN has critical problems; for example, it uses local information for upsampling. Hence, pixels belonging to large objects are classified as different objects, and small objects are labeled as background [21]. Their accuracy depends on the size of the available training sets and the considered networks' size. Moreover, they are pretty slow because the network has to run separately for each patch [22]. After FCNs, encoder-decoder architecture came, Like SegNet [23], U-net [22], and Deconvolution Network [21]. This architecture has been using techniques that make it powerful compared to FCN. These techniques are tried to gradually make pixel-wise labels from feature maps to prevent losing data, using an up-sampling method instead of pooling layers in FCN and inserting skip connections from the encoder to the decoder to recover data better. By using these techniques, it achieves good accuracy on a small dataset for training [22]. In this approach, fully connected layers can be omitted to save memory.

After fully connected networks, dilated or atrous convolutions have been proposed. Because pooling layers decrease the resolution of images, dilated came. In this method, the size of the filter increase with the insertion of zeros. So, it lets us enlarge the filter size without increasing the number of parameters. It means atrous convolution allows networks to consider each pixel and its neighbors to extract features without increasing the number of parameters [9], [24], [25]. The method of [26] is very fast and lets us capture contextual information and denser feature map [4], [9]. Dilated conventional layers are used in well-known architectures, such as DeepLabv1 [26], DeepLabv2 [24], and DeepLabv3 [27]. ParseNet [28] is image-based and uses contextual information, which contrasts with FCN, which is pixel-based and does not utilize global details. ParseNet does an up-sample operation on the feature map directly and combines this with global information extracted with global pooling using L2 normalization instead of convolution layers.

In some architectures, features are extracted from different scales of the image, known as pyramid methods. Pyramid methods are based on the fact that up-sampling layers cannot preserve low-level features, and the information of down-sampling operations, like boundaries, will be missed. Hence, the output is not accurate enough. Networks have tried to preserve low-level features to prevent low accuracy. Dilated convolutional layers in DeepLabv1 try to keep contextual information. However, they have to run on large numbers and high dimensions of the feature map, which is computationally expensive.

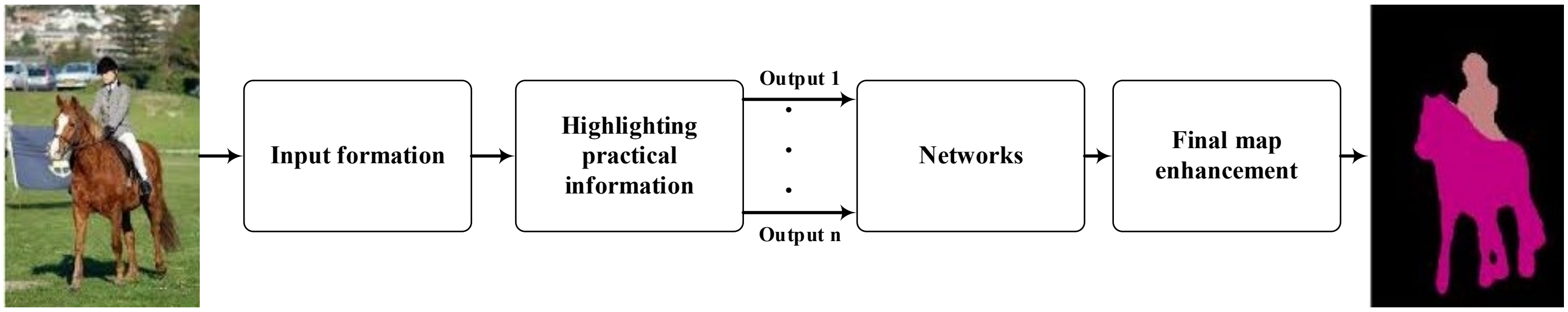

Fig 1. Proposed abstraction model for semantic segmentation methods

On the other hand, dilated convolution leads to the loss of essential details. In RefineNet, multiple connections from different resolutions are used, and it tries to fuse low-level features with high-level features to prevent the disadvantage of down-sampling operation [29]. Pyramid Scene Parsing Network (PSPNet) tries to consider global scene categories and information for classifying each pixel [3]. In this model, pyramid pooling modules are used to prevent information loss between sub-regions after preparing a feature map with a dilated network strategy in ResNet [30]. As it was mentioned before, DeepLab v2 [24] and v3 [27] first use atrous convolution to extract feature maps; second, they use Atrous Spatial Pyramid Pooling (ASPP) to use global information. Unlike DeepLab v3, DeepLab v2 employs CRF in final map enhancement. DeepLabv3 uses an image-level view like ParseNet instead of a pixel-level to improve accuracy and use global information. In some networks, features are extracted from spatial regions. Mask RCNN [2] is a region-based method and is used for semantic segmentation. This method uses Faster-RCNN [31]. Also, Mask RCNN uses Region Proposal Network (RPN) to extract the Region of Interest (ROI). It is modified with FCN to predict the semantic segmentation task's mask and boundary boxes for each ROI. Instead of ROI Pooling in Faster-RCNN, it uses ROIAlign to extract small feature maps from each ROI to preserve location better. Some of these networks could be used for output enhancement. In semantic segmentation, Conditional Random Fields (CRF) are usually used for final map enhancement. CRFs have high computational complexity, slow, and hard to optimize. But recently, some methods faster than CRF, like convolutional CRFs, have been proposed.

Besides these FCNs, Recurrent Neural Networks (RNN) could be used for semantic segmentation. These methods help us to find dependencies in semantic segmentation tasks [32],[33]. Furthermore, RNN-based methods try to smooth their predicted labels and use local and global features, meaning RNNs extract contextual information.

As mentioned before, different approaches have been introduced, but no one has published a general structure for semantic segmentation. With the help of an abstract framework, we can analyze different approaches and see each abstracted part's effect on the method's performance. For this purpose, in this paper, a general framework for semantic segmentation is proposed.

The rest of this paper is organized as follows. First, section II describes different blocks of the framework. Then, in each sub-section, one sub-block is explained. Finally, in section III, the conclusion will be described, and a comparison of the mentioned methods will be made.

## II. Proposed Framework

In this study, we are going to demonstrate the abstract view for semantic segmentation and show a block diagram for this task that other papers have followed. The block diagram of this framework is shown in Fig. 1. In this abstract framework, the input image passes through a pipeline with several blocks, (1) input formation, (2) highlighting practical information, (3) networks, and (4) Final map enhancement. According to its application, input formation, highlighting useful information, and the final map enhancement could be omitted. The network is used to label and classify every pixel of an input image. In the following, each step of this framework will be explained.

### A. *Input Formation*

This block attempts to enhance input images. This can be done by augmenting images via the augmentation block or improving image quality via the image enhancement block. Two sub-blocks, augmentation and image enhancement, are described below.

#### 1) *Augmentation*

Several papers have used augmentation in different situations to increase input data and prevent overfitting neural networks. This process is useful in most cases, especially when there are not enough data or it is hard to collect data and prepare ground truth in situations like medical datasets [9], [8]. With the help of augmentation, the number of data multiplies through simple techniques, like scaling [2], cropping, rotating [1], [33], re-sizing, flipping [34], mirroring, and jittering [17], [5], shifting [22], extra annotations [26], [24], [27], [28], [29], [35] and inverse detectors [3], [21]. In [5], Gaussian blur is used. This method enhances images. But if both enhanced and non-enhanced images are used, it can augment and increase data. Other complex techniques like GAN have been used for augmentation, too [36].

#### 2) *Image enhancement*

Image enhancement improves essential features and reduces input noises and inconsistencies [37]. There are different methods for pre-processing and image enhancement like re-sizing input image, histogram equalization [38], [8], normalization [38], [9], Gaussian blur [5], Gaussian distribution, and random displacement vectors [22].

### B. Highlighting Practical Information

In this block, we extract practical information to help the main network train more efficiently or select interest zones for sending to the primary networks. The extracted information and features highlight image contents, pixels, and other properties. These properties help networks work better or pay attention to specific zones. Ultimately, we combine the output of different levels and prepare them to feed into the network. The block extracts the zone of interest (ZOI) and prepares auxiliary information.

#### 1) ZOI extraction

The ZOI extraction method tries to determine some specific zones and then feeds them into the network [39]. In some cases, we could do it manually [40]. However, some methods do it automatically with trained networks. For example, after feature extraction in this network, another network is used to find ROI [2]. After feature extraction in the primary network, extracting ZOI gives us excellent accuracy and speed because it lets the network work on features, not raw images. However, using sliding windows on a raw image is a traditional method and computationally expensive.

#### 2) Auxiliary information preparation

Semantic segmentation can be trained easier if the primary network knows about objects' locations, the type of scene, and objects in the image. In [41], CRF tries to understand the kind of image scene. Hence, this approach can improve network output since it narrows down the classes of semantic segmentation. Some papers like [1] use auxiliary information, like edges, that may weaken gradually in deep networks. Hence, extra features preserve the deep networks from forgetting low-level information and help the network to classify objects as more important. Also, in this block, one can use different input data channels like the digital elevation model (DEM) [1]. Then parallel networks train for every channel in the network block. The auxiliary information is combined with input images in this block and sent to the network block. As mentioned, one can omit this block and train parallel blocks for different kinds of data prepared in the previous phase [1].

### C. Networks

Networks block attempts to extract different features and then use them for classification. The output is a pixel-level classified image, and sometimes extra features are extracted to improve output. Additional features have been used to employ in final map enhancement to refine the network's outputs, as is shown in Fig. 2. In the proposed block, one can see several different networks that may vary in their input, type of backbone, trained dataset, or type of extracted features.

#### 1) Network architecture and auxiliary feature selection

In this block, networks' architecture, the number of networks, their input, and extra features needed in the final map enhancement have to choose. The following two sub-blocks will discuss which networks are suitable for every target and essential parameters for network selection.

#### 2) Feature extraction and classification

Networks are chosen based on goals, limitations, and databases. For example, to prevent losing low-level features in [22], [23], [21], [8], [9], and [41], skip connections are used to improve accuracy and boundary detection by connecting features from different levels. Also, skip connections let us prevent vanishing gradient [42]. If one wants to use global features in classification, there are various ways like dilated convolution, pyramid models, RefineNet, ParseNet, and RNNs. They attempt to fuse global features with local features. If one faces time and memory limitation problems, one should use a light network like SegNet [23] or depth-wise separable convolution [4]. One can use different networks trained with different datasets if one's goal is to accomplish higher accuracy without having memory or time limitations. Hence, each network may cover another network's problem and achieve more accuracy [1].

#### 3) Auxiliary feature extraction

As mentioned before, some networks can extract special features to help the final map enhancement block for smoothing output, noise illumination, and boundary optimization. For example, some methods extract superpixels [43] and edges [3]. These networks try to extract some features to improve the probability of edges or reduce noise probability. Low-level features extracted in semantic segmentation networks can help these methods to get extra information like edges. After combining different auxiliary features with classification networks output, the result is sent to the final map enhancement block. There is no special method for a combination of results. Conv1*1 [27], [1], L2 norm [28], sum, and different methods are used for the combination [43].

### D. Final Map Enhancement

Final map enhancement attempts to improve performance and accuracy with mathematical algorithms or popular networks. It also has been used to fix mistakes and may be one of the essential blocks in semantic segmentation. Different methods are employed for final map enhancement and refining output results. Final map enhancement methods are divided into two groups. The first group tries to improve boundaries like domain transform (DT) [3] and boundary optimizer [43]. The second group is boundary optimizer, image smoothing, and pixel-wise classification. In addition, it is employed to use global and contextual information like CRF and MRF-based methods [23], [5], [1], [9], [24], [43], [23], [38], [26], [28], [29], [21]. The Second group is wildly used in semantic segmentation, and there are different derivatives based on CRF and MRF, like DenseCRF [44]. The disadvantage of these methods is that they are highly computational and time-consuming; hence, researchers use alternative methods like BNF [45].

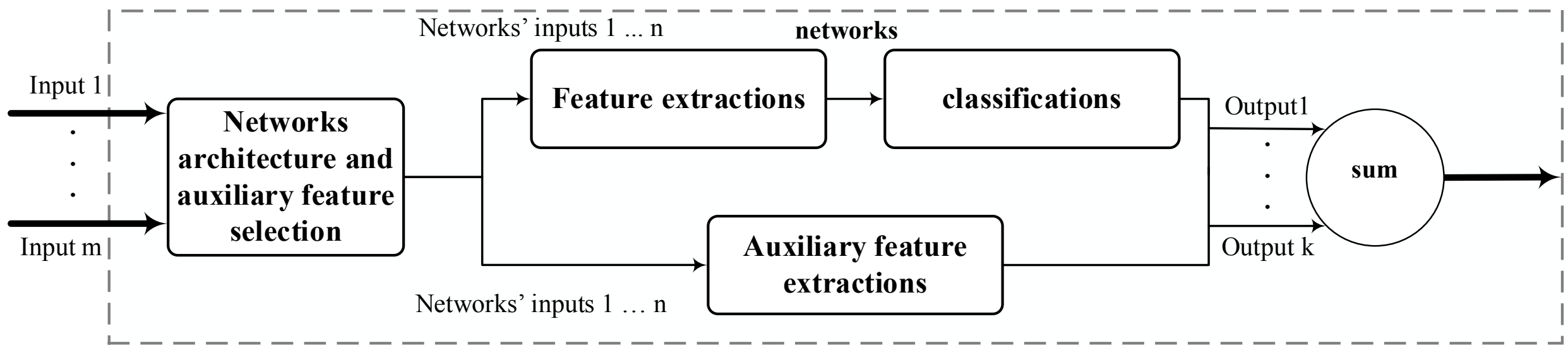

Fig 2. The network part of the framework.

## III. CONCLUSION

In this paper, we proposed a global platform for semantic segmentation. This framework could divide most of the presented semantic segmentation methods into four basic blocks. According to the presented methods, each block and subblock have an essential role in this process.

As presented, the Input formation block tries to prevent overfitting by data augmentation and helps the network train better. Image enhancement methods are also used to augment data. But it is employed to improve image quality and resolution too. Highlighting practical information has been rarely used because ZOI extraction does not work efficiently with raw images, but networks are potent these days. Therefore, selecting ZOI from the feature map, not the raw input image, is better.

On the other hand, extra features cannot help deep networks a lot. Using deep networks with unique tricks that are mentioned is more beneficial. For example, using a global point of view with pixel-wise information in the network blocks yields higher efficiency. Also, not losing low-level features in up-sampling and employing them directly in the decoder process helps a network to become highly efficient.

On the other hand, some networks, like Deeplabs, use pyramid design to improve accuracy. Parallel networks with different input data and trained datasets let the network consider various aspects of the input image. However, this trick is costly and not recommended when limited resources are used. The significance of the final map enhancement block is undeniable. It tries to reduce noise and smooth output results and attempts to enhance boundaries. Extracting extra features from the block, such as boundary information, leads to smooth and noise-free output. CRF and its Derivatives are computationally expensive but are well-known methods in final map enhancement.

Many presented methods are considered in this paper. In these methods, different datasets are used to evaluate their works. However, Pascal Voc 2012 dataset is more common compared to others. In TABLE [1], the results of papers that use this dataset are shown. By comparing the accuracy and the utilized blocks in these methods, it is deduced that algorithms that use features in different scales could reach higher accuracy. This table shows that strategies that use DeepLabV3, V3+, PSPNet, and RefineNet benefit from multiscale features via pyramid networks and dilated convolutions. Hence, these reach better accuracy compared to others.

TABLE I. COMPARING DIFFERENT SEMANTIC SEGMENTATION ALGORITHEMS

| | **Input formation** | **Networks** | **Final map enhancement** | | |
|---|---|---|---|---|---|
| **paper** | Augmentation & Image enhancement | Feature extraction & classification | Final map enhancement | Data base | accuracy |
| [4] | | DeepLabv3+, Xception, Depth wise separable convolution | - | PASCAL VOC 2012 | 89.0% |
| [27] | extra annotations | DeepLab3 | - | PASCAL VOC 2012 | 85.7% |
| [5] | Mirroring, resizing, rotating, Gaussian blur | PSP | - | PASCAL VOC 2012 | 85.4% |
| [34] | Flipping and scaling | Proposed network (Pyramid Attention Network (PAN)) | - | PASCAL VOC 2012 | 84.0% |
| [29] | Scaling, random cropping, horizontal flipping | RefineNet | CRF | PASCAL VOC 2012 | 83.4 |
| [24] | extra annotations | DeepLabv2 | CRF | PASCAL VOC-2012 | 79.7% |
| [43] | | FCN(VGG+FCN32) + SLIC (super pixel) | Boundary optimizer, CRF | VOC 2012 | 74.5% |
| [21] | Inverse Detectors | Proposed encoder-decoder network | CRF | PASCAL VOC 2012 | 72.5% |
| [26] | extra annotations | DeepLabv1 | CRF | PASCAL VOC-2012 | 71.6% |
| [35] | flip, rotation, resize, jittering color | Convolutional CRFs | - | Pascal VOC 2012 | 71.23% |
| [3] | Inverse Detectors | DeepLab + EdgeNet (Edge prediction) | Domain transform | PASCAL VOC 2012 | 71.2%. |
| [40] | | convolutional neural networks | - | PASCAL VOC 2012 | 69.6% |
| [17] | mirroring and jittering | FCN | - | PASCAL VOC | 62.2% |